\documentclass[sigconf]{acmart}

\usepackage{booktabs}
\usepackage{multirow}
\usepackage{graphicx}
\usepackage{caption}
\usepackage{subcaption}
\usepackage{enumitem}
\usepackage{amsmath}
\usepackage{hyperref}
\usepackage{xcolor}

\acmConference[UISE '26]{1st International Workshop on User Interface and Experience for Software Engineering (UISE 2026)}{April 2026}{Rio De Janeiro, Brazil}

\title{Bias Beneath the Tone: Empirical Characterisation of Tone Bias in LLM-Driven UX Systems}

\author{Heet Bodara}
\affiliation{%
  \institution{Faculty of Information Technology, Monash University}
  \city{Melbourne}
  \state{VIC}
  \country{Australia}}
\email{hbod0002@student.monash.edu}

\author{Md Masum Mushfiq}
\affiliation{%
  \institution{Faculty of Information Technology, Monash University}
  \city{Melbourne}
  \state{VIC}
  \country{Australia}}
\email{md.mushfiq@monash.edu}

\author{Isma Farah Siddiqui}
\affiliation{%
  \institution{Faculty of Information Technology, Monash University}
  \city{Melbourne}
  \state{VIC}
  \country{Australia}}
\email{ismafarah.siddiqui@monash.edu}

\begin{document}

\begin{abstract}
Large Language Models are increasingly used in conversational systems such as digital Personal Assistants, shaping how people interact with technology through language. While their responses often sound fluent and natural, they can also carry subtle tone biases such as sounding overly polite, cheerful, or cautious even when neutrality is expected. These tendencies can influence how users perceive trust, empathy, and fairness in dialogue. In this study, we explore tone bias as a hidden behavioural trait of LLMs. The novelty of this research lies in the integration of controllable LLM-based dialogue synthesis with tone classification models, enabling robust and ethical emotion recognition in PA interactions. We created two synthetic dialogue datasets: one generated from neutral prompts and another explicitly guided to produce positive or negative tones. Surprisingly, even the neutral set showed consistent tonal skew, suggesting that bias may stem from the model’s underlying conversational style. Using weak supervision through a pretrained DistilBERT model, we labelled tones and trained several classifiers to detect these patterns. Ensemble models achieved macro-F1 scores up to 0.92, showing that tone bias is systematic, measurable, and relevant to designing fair and trustworthy conversational AI.
\end{abstract}




\maketitle

\section{Introduction}
User interfaces (UIs) are no longer confined to screens and buttons; they increasingly take the form of intelligent, conversational agents that understand and respond to natural language. Among these, \textit{digital personal assistants (PAs)} such as Siri, Alexa, Cortana, and Google Assistant have become part of everyday life, helping users retrieve information, manage schedules, and perform tasks through dialogue. As these systems evolve, their tone, phrasing, and conversational style have become central to the overall \textit{user experience (UX)}. Tone affects how users perceive trust, empathy, and fairness in their interactions, qualities that directly shape the usability and acceptability of intelligent interfaces.

With recent advances in Large Language Models (LLMs), it is now possible to simulate rich, human-like dialogue at scale and to prototype conversational UIs without collecting real user data. However, this capability introduces a subtle but important concern: LLMs themselves may inject or amplify \textit{tonal bias} in the interfaces they power. If an assistant consistently responds in an overly positive, dismissive, or hedged tone, it can distort users’ perception of reliability, politeness, or confidence, affecting both usability and trust. Addressing such stylistic bias, therefore, becomes not just a linguistic challenge but a core aspect of UI/UX engineering.

To examine this phenomenon, we generated two synthetic datasets of user–assistant dialogues using several state-of-the-art LLMs. One set contained tone-neutral conversations without explicit instructions, while the other was tone-conditioned through prompts specifying positive or negative sentiment. Although the conditioned data behaved as expected, even the neutral set displayed consistent tonal tendencies, suggesting that bias may arise from the model’s underlying conversational style rather than user input. 

\textbf{Research questions.} Guided by this motivation, our study investigates:
\begin{itemize}
    \item \textbf{RQ1:} How can LLMs be leveraged to generate realistic, tone-diverse synthetic datasets that emulate real human–assistant dialogues?
    \item \textbf{RQ2:} How effectively can machine-learning, deep-learning, and ensemble models trained on such data classify and generalise LLM-induced tones across varied contexts?
\end{itemize}

\textbf{Approach and findings.} To address \textbf{RQ1}, we generated two complementary synthetic datasets of user–assistant dialogues using several state-of-the-art LLMs. The first captured tone-neutral exchanges without any emotional instructions, while the second was explicitly tone-conditioned through prompts specifying positive or negative sentiment. This controllable generation process enabled scalable and reproducible simulation of tone-diverse interactions. To address \textbf{RQ2}, we applied weak supervision with a pretrained DistilBERT model to label tone and trained a suite of classifiers, from lightweight TF–IDF and linear models to neural memory networks on both datasets. Stricter labeling thresholds ($\tau = 0.85$) produced clearer tone separation and higher macro-F1 scores (0.84–0.92), while inclusive thresholds ($\tau = 0.60$) surfaced borderline cases. The ensemble of Logistic Regression and Linear SVM achieved the most stable cross-dataset performance, while neural variants captured subtler contrastive tones such as sarcasm and hedging.

Overall, the results reveal that tonal bias in LLM-driven dialogues is systematic and measurable. Even tone-neutral prompts yield consistently polite or positive framings, suggesting that stylistic bias emerges from model behaviour rather than user intent. The study’s novelty lies in combining controllable LLM-based dialogue synthesis with interpretable tone classification, offering a lightweight yet effective diagnostic pipeline for examining tone bias. By bridging bias analysis with user-experience evaluation, this work advances the design of conversational systems that are more transparent, fair, and trustworthy in both content and delivery.

\section{Background and Related Work}
Emotion identification in dialogues has long been a central topic in natural language understanding, supporting work in areas such as opinion mining, mental health monitoring, customer support, and human-computer interaction. Much of this progress stems from the growing realisation that effective communication between people and machines depends not only on literal accuracy but also on emotional sensitivity. As voice assistants like Alexa, Siri, Cortana, and Gemini have entered daily life, expectations around empathy and tone have increased sharply. These systems are now judged as much by \emph{how} they respond as by \emph{what} they say. From a UX perspective, tone and emotional awareness have become essential to trust and comfort in everyday interaction.

Earlier studies in dialogue emotion analysis (DEA) concentrated mainly on detecting human emotions in structured or scripted corpora such as movie dialogues, debate transcripts, or controlled conversation datasets. Gan et al.~\cite{gan2024survey} provide an extensive survey covering work from 2017 to 2024, mapping methodological trends and open ethical questions. They identify three broad modelling paradigms that dominate the field:
\begin{itemize}
    \item \textbf{RNN-based approaches}, which follow sequential dynamics to track speaker states~\cite{Majumder_Poria_Hazarika_Mihalcea_Gelbukh_Cambria_2019,Li_Jiang_2024,WEN2023123,LI202273};
    \item \textbf{GNN-based approaches}, capturing relations between speakers and utterances through graph structures~\cite{ghosal-etal-2019-dialoguegcn,ishiwatari-etal-2020-relation,TU_IEEE_TAC_2022,Yang_2024_IEEE_TAC};
    \item \textbf{Transformer-based methods}, which rely on attention and large pre-trained language models for contextual emotion representation~\cite{shen2024multimodal,HUI_10.1109/TMM.2023.3271019,li-etal-2022-emocaps,zhu-etal-2021-topic}.
\end{itemize}

These approaches collectively advanced context-sensitive emotion modelling, yet their real-world transfer remains limited. Domain shifts, sparse emotional data, and latency constraints still make it hard to deploy them in live personal assistants.

To improve realism, researchers have explored multimodal and adaptive systems. Sindhu et al.~\cite{Sindhu_2024} describe a two-stage chatbot that fuses text and audio for empathetic replies. Abinaya et al.~\cite{abinaya2024} propose TEBC-Net, blending BERT text encoders with CNN-based facial analysis to align responses with perceived mood. Kovacevic et al.~\cite{kovacevic2024} contribute a dataset of genuine human–chatbot exchanges, showing that personalisation reduces domain gaps. In parallel, Zheng et al.~\cite{zheng2025chi} build \emph{ChatLab}, letting users customise LLM-powered support bots; their emphasis lies on user experience rather than tone recognition. Pias et al.~\cite{pias2024cui} analyse how the assistant’s tone, apparent age, and gender sway engagement and purchase intent, interesting from a UX lens, though not a direct study of tone detection.

Recently, attention has turned from reading emotions in users to examining \emph{how models themselves express tone}. Bardol~\cite{bardol2025emotional} finds that simply altering the emotional framing of a prompt can shift the polarity of an LLM’s replies, even when content remains constant. Dobariya and Kumar~\cite{dobariya2025mind} show that polite prompts elicit more accurate and confident responses, highlighting sensitivity to pragmatic framing. Laurito et al.~\cite{laurito2025aiai} observe that models prefer text written in a “machine-like’’ style, effectively displaying bias toward their own generative tone. Vinay et al.~\cite{vinay2025emotional} link emotional prompting to a rise in misinformation, suggesting an interaction between affect and factual reliability. Finally, Gallegos et al.~\cite{gallegos2024biasfairnesslargelanguage} survey bias and fairness in LLMs, noting that stylistic and tonal patterns form a distinct, under-studied dimension of bias.

Taken together, these works point to a shift: tone and framing are no longer just linguistic artefacts but measurable design factors influencing user perception. Our work follows this direction but asks a different question, \emph{how do these tonal biases surface inside the conversational agents themselves}? We treat the LLM both as a data generator and as a subject of evaluation, tracing its default tone even when neutrality is requested. This perspective connects emotion-recognition research with broader issues of fairness and reliability in intelligent user interfaces, aligning with the UISE goal of creating user-centred, transparent, and trustworthy interactive systems.


\section{Methodology and Implementation}
\begin{figure}[!h]
    \centering
    \includegraphics[width=\columnwidth]{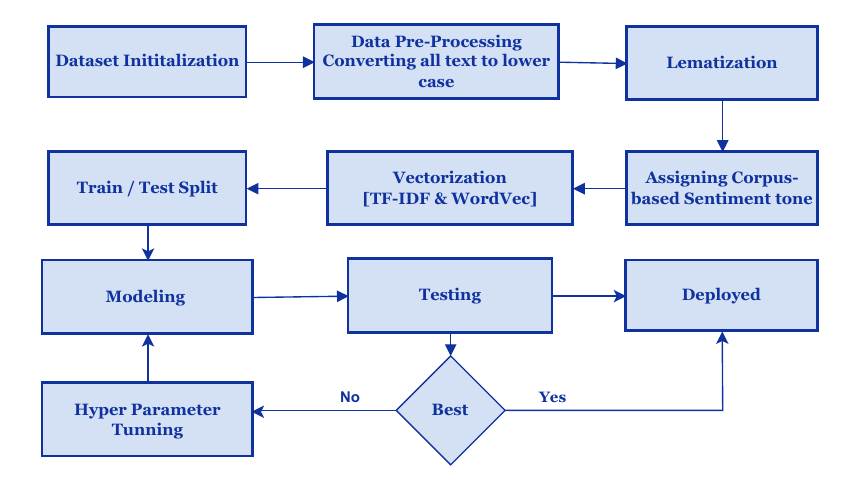}
    \caption{Overview of the workflow.}
    \label{fig:methodology}
\end{figure}

 Our workflow, illustrated in Figure~\ref{fig:methodology}, follows a standard yet lightweight NLP pipeline designed to diagnose tonal bias in assistant-style responses. We begin by constructing two synthetic datasets of user--assistant question-answer pairs. The first dataset contains general, tone-neutral conversations generated using multiple large language models (LLMs), including ChatGPT, Google Gemma, and Open Hugging Face variants. The second extends this with tone-conditioned prompts (positive and negative), ensuring a balanced sample distribution and richer lexical variety.

After dataset initialization, all responses are normalized and lemmatized to reduce lexical sparsity. Text is converted to lowercase, extraneous symbols are cleaned, and samples outside the range of three to two hundred tokens are removed. Each assistant reply is then assigned a sentiment tone using a pretrained \texttt{DistilBERT} model fine-tuned on SST-2, which provides weak labels for positive, negative, or neutral tone depending on confidence thresholds. This automated annotation allows scalable labeling while limiting manual bias.

For feature representation, we employ two complementary encodings: a sparse term-frequency inverse-document-frequency (TF-IDF) model capturing lexical cues or a dense Word2Vec embedding capturing contextual similarity. The corpus is divided into training and testing sets in an 80/20 ratio, maintaining tone balance and topic stratification. We then train several lightweight classifiers, Multinomial Naïve Bayes, Logistic Regression, and Linear SVM, and compare them with an ensemble combining Logistic Regression and SVM via probability averaging. Exploratory neural models, Neural Semantic Encoder (NSE) and Dynamic Memory Network (DMN), were also examined to evaluate their sensitivity to mixed or contrastive tones~\cite{app13074550,dimalanta2025icit,TAHA2024100664,munkhdalai-yu-2017-neural-semantic,ZHANG201836}. MNB, LR, and Linear SVM are competitive, interpretable baselines for short texts that run well on modest hardware; an LR+SVM ensemble further stabilizes performance by combining calibrated probabilities with max-margin decisions, while NSE/DMN probes whether memory/attention mechanisms expose subtle bias that bag-of-words may miss (e.g., hedging, contrastive framing, sarcasm).

In the ensemble, we used two methods: \textit{soft voting} and \textit{stacking}.
\begin{itemize}
    \item \textbf{Soft voting (probability averaging)}: 
    
    For class $c \in \{-1, +1\}$ and base models $k = 1, \dots, K$ with calibrated posteriors $p_k(c \mid \mathbf{x})$, the ensemble prediction is computed as:
    \[
    p_{\text{ens}}(c \mid \mathbf{x}) = \sum_{k=1}^{K} w_k \, p_k(c \mid \mathbf{x}), 
    \quad 
    \sum_{k=1}^{K} w_k = 1, \; w_k \ge 0,
    \]

    \[
    \hat{y} = \arg\max_{c} \; p_{\text{ens}}(c \mid \mathbf{x}).
    \]

    \item \textbf{Stacking (logistic combiner)}: 
    
    Let $\mathbf{z} = [p_1(+1 \mid \mathbf{x}), \dots, p_K(+1 \mid \mathbf{x})]$ be the vector of base model posteriors. 
    A logistic meta-model is trained on validation folds to learn:
    \[
    p_{\text{ens}}(+1 \mid \mathbf{x}) = \sigma(\beta_0 + \boldsymbol{\beta}^{\top}_\mathbf{z}),
    \quad
    \hat{y} = \mathbf{1}\{p_{\text{ens}}(+1 \mid \mathbf{x}) \ge \tau\}.
    \]
\end{itemize}

    
    

    

Hyperparameter tuning was performed manually within compact ranges (e.g., $\alpha = 0.1$--$1.0$ for NB, $C = 0.1$--$3$ for LR/SVM). Training and evaluation were conducted in Google Colab using Python~3.x, \texttt{scikit-learn}, and \texttt{Hugging~Face~Transformers}. Classical models executed efficiently on CPU, while DistilBERT labeling and neural variants used a single T4 GPU when available. 

Overall, this pipeline (Figure~\ref{fig:methodology}) provides a reproducible foundation for examining tone bias in LLM-generated data. It emphasizes interpretability and diagnostic precision rather than complex model design, allowing the analysis to focus on where bias appears and how different model families detect it.

\section{Results and Discussion}
Our analysis focuses on how effectively different model families detect tonal bias in LLM-generated assistant responses. Experiments were conducted on two complementary datasets: one consisting of general tone-neutral dialogues and another containing tone-conditioned prompts. Performance was evaluated across two confidence thresholds ($\tau=0.60$ and $\tau=0.85$) applied during the DistilBERT-based weak labeling stage. These thresholds control how inclusive or conservative the sentiment assignments are.

\begin{figure}[t]
  \centering
  \begin{subfigure}[b]{0.48\columnwidth}
    \includegraphics[width=\linewidth]{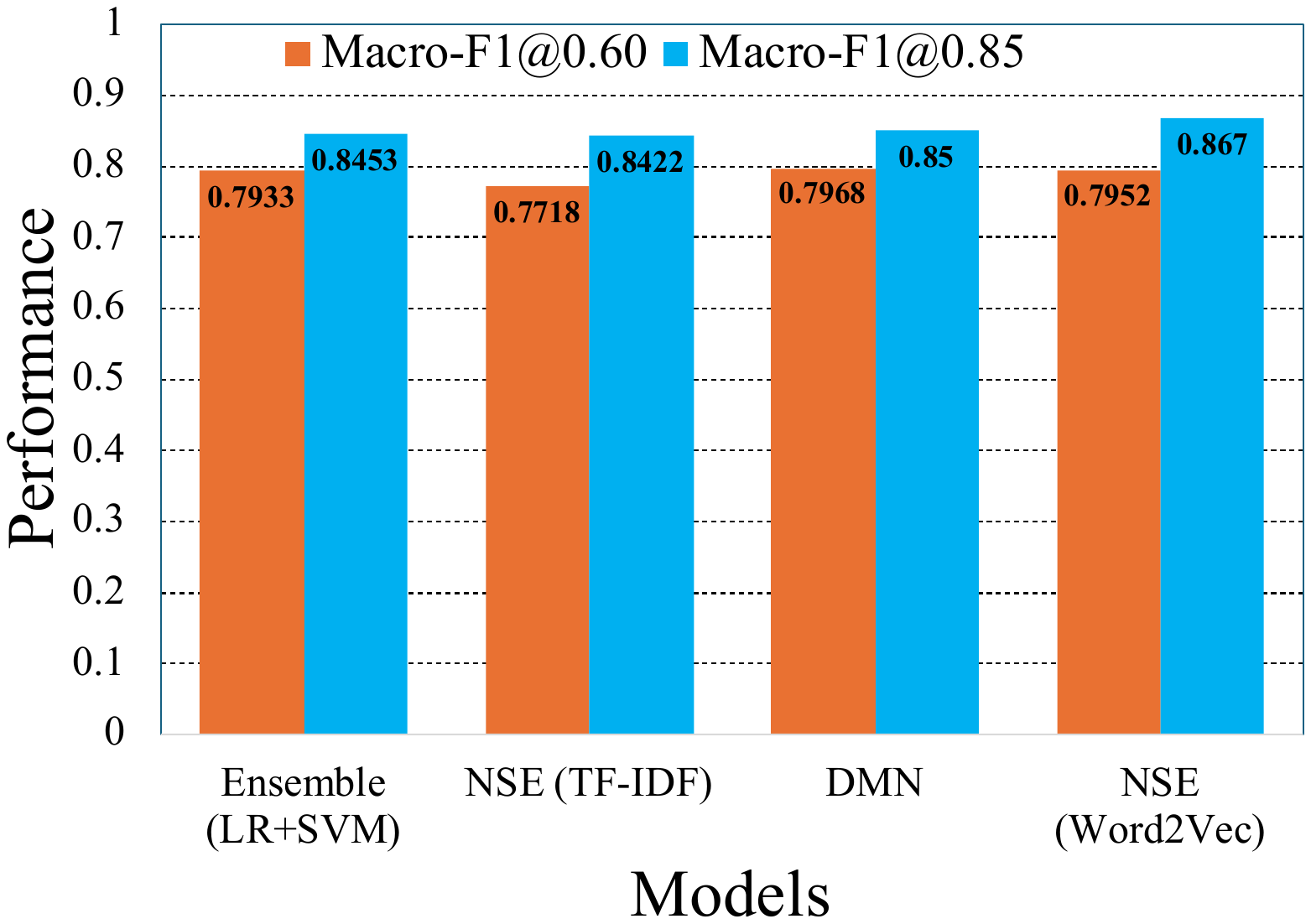}
    \caption{Results for Dataset A.}
    \label{fig:f1_threshold}
  \end{subfigure}
  \hfill
  \begin{subfigure}[b]{0.48\columnwidth}
    \includegraphics[width=\linewidth]{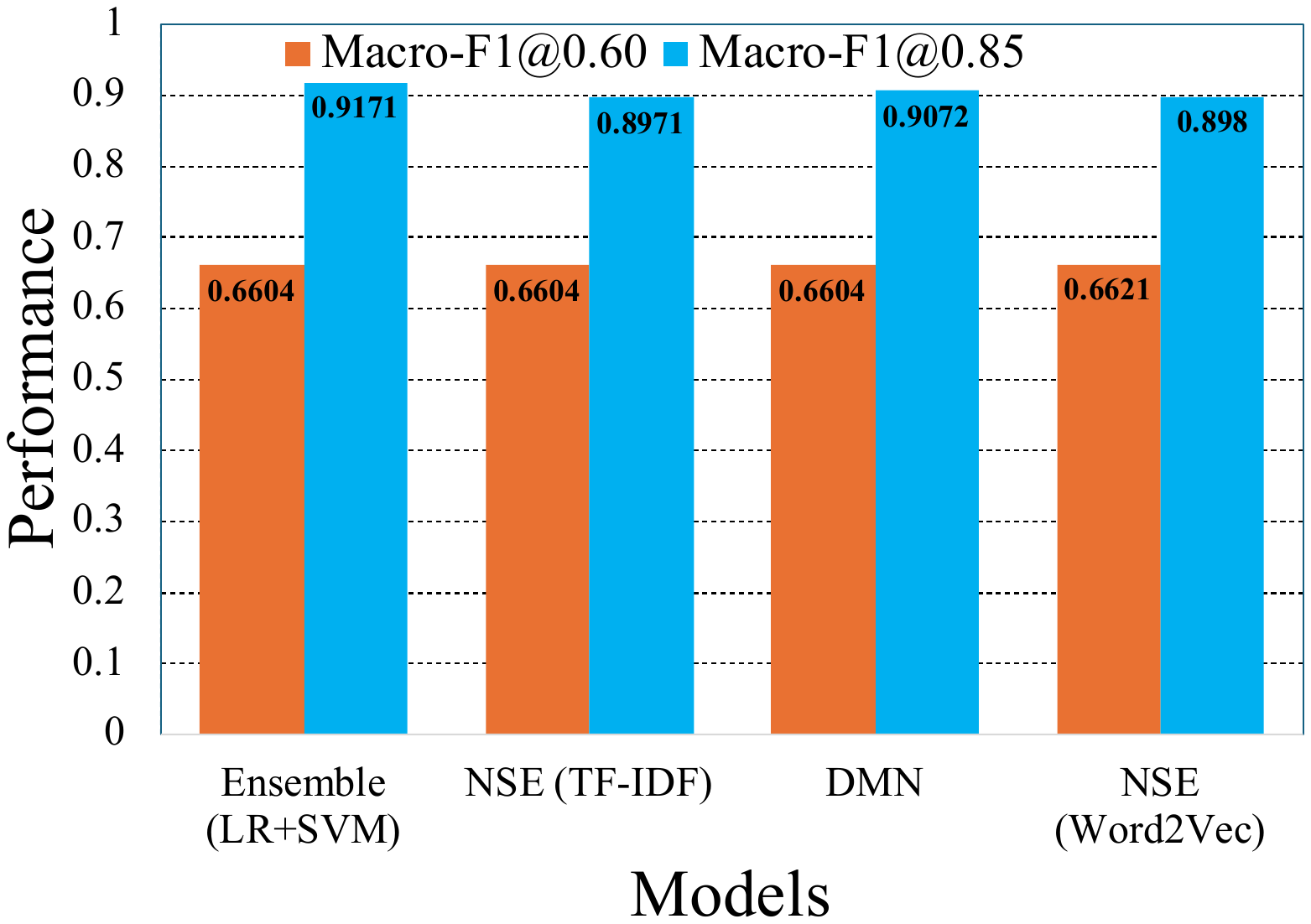}
    \caption{Results for Dataset B.}
    \label{fig:model_comparison}
  \end{subfigure}
  \caption{Macro-F1 performance of different models across tone-classification tasks.}
  \label{fig:ensemble_results}
\end{figure}

At the lower threshold ($\tau=0.60$), the models captured more borderline or ambiguous cases but at the cost of reduced precision. Accuracy remained high (above 0.96), yet macro-F1 scores dropped to the 0.66-0.80 range, indicating frequent confusion between neutral and mild positive tones. Increasing the threshold to $\tau=0.85$ produced clearer separations between classes, yielding macro-F1 values between 0.84 and 0.92 across both datasets. As illustrated in Figure~\ref{fig:f1_threshold}, stricter thresholds consistently improved macro-F1 and reduced label noise, confirming that tone bias becomes more apparent when ambiguous samples are excluded.

Among classifiers, the ensemble of Logistic Regression and Linear SVM achieved the best overall balance between precision and recall, followed closely by the DMN and NSE. Figure~\ref{fig:model_comparison} compares results across the two datasets, showing that model rankings remain stable and that linear TF-IDF models perform surprisingly well despite their simplicity. The ensemble provides modest yet consistent gains by averaging decision boundaries, while neural variants add small improvements on nuanced cases.


Qualitatively, we observed that the most error-prone cases were subtly framed replies where lexical positivity contrasted with negative implications (\textit{e.g.}, polite disagreement or cautionary advice). Sarcastic or hedged responses also challenged linear models, as these cues are often expressed through punctuation or rhythm rather than explicit word choice. Neural models captured some of these patterns through memory and attention mechanisms but remained sensitive to label noise. Domain-wise, factual and short-form prompts (such as productivity or finance) showed higher precision, while advice- or opinion-driven categories (health, news) exhibited stronger bias and lower recall.

Overall, the results highlight that tonal bias can be systematically identified using interpretable models trained on weakly labeled data. Adjusting labeling thresholds provides a practical means to balance inclusiveness against diagnostic confidence, allowing us to quantify not only \textit{where} bias appears in assistant-style responses but also \textit{how confidently} different model types detect it.



\section{Conclusion and Future Work}
This study examined tonal bias as a diagnostic vulnerability in LLM-generated voice-assistant responses. By constructing two synthetic datasets, tone-neutral and tone-conditioned, and applying weak supervision through DistilBERT sentiment labeling, we showed that even unprompted LLM outputs exhibit systematic tonal tendencies. Classical linear models trained on TF-IDF representations proved both effective and interpretable, while ensemble and neural variants offered incremental gains, particularly in detecting nuanced or contrastive tones. Our findings indicate that tonal bias is not only quantifiable but also threshold-dependent: inclusive labeling captures subtle variations, whereas conservative labeling isolates clearer polarity boundaries. This trade-off provides a practical diagnostic tool for assessing bias direction and severity across assistant responses. 

In the future, we plan to replace weak supervision with an expert-annotated bias rubric and a calibrated gold-standard dataset to enable finer-grained, context-aware detection. We will incorporate contextual encoders and cross-validation to capture subtle tones such as neutrality and hedging, and extend the analysis to multilingual and counterfactual dialogue data. To ensure ethical and practical validity, we will evaluate the framework on real personal-assistant interactions and introduce adversarial and transfer tests. Finally, we aim to develop interpretable auditing tools and a diagnose → intervene → remeasure dashboard for transparent, continuous bias monitoring in conversational AI.

\bibliographystyle{unsrt}
\bibliography{references}

\end{document}